\documentclass{article}
\pdfoutput=1
\usepackage{PRIMEarxiv}
\usepackage[utf8]{inputenc} 
\usepackage[T1]{fontenc}    
\usepackage{xcolor}
\definecolor{InvisibleGreen}{rgb}{0.90, 0.90, 0.89}
\definecolor{InvisibleBlue}{rgb}{0.95, 0.85, 0.92}

\definecolor{MediumRed}{rgb}{0.99, 0.145, 0.145}
\definecolor{MediumGreen}{rgb}{0.17, 0.67, 0.26}
\definecolor{MediumBlue}{rgb}{0.35, 0.315, 0.55}

\definecolor{DarkBlue}{rgb}{0.105, 0.15, 0.30} 
\definecolor{DarkRed}{rgb}{0.60, 0.105, 0.15} 

\definecolor{roseRed}{HTML}{e20047}
\definecolor{coolGray}{HTML}{474747} 
\definecolor{lightbox}{HTML}{e9ecef}
\definecolor{peachPink}{HTML}{ff6f61}
\definecolor{coralRed}{HTML}{ff412e}
\definecolor{softLilac}{HTML}{c1a7d1}
\definecolor{deepBlue}{HTML}{003366}
\definecolor{pineGreen}{HTML}{00593B}
\definecolor{racingGreen}{HTML}{074B33}

\definecolor{cherryRed}{HTML}{d50032}  
\definecolor{burgundy}{HTML}{800020}  
\definecolor{salmonPink}{HTML}{ff91a4}  
\definecolor{crimson}{HTML}{dc143c}  
\definecolor{darkRose}{HTML}{c71585}  
\definecolor{carmine}{HTML}{960018}  
\definecolor{indianRed}{HTML}{cd5c5c}  
\definecolor{raspberryRed}{HTML}{e30b5c}  
\definecolor{fuchsia}{HTML}{ff00ff}  
\definecolor{rosewood}{HTML}{65000b}  
\definecolor{tomatoRed}{HTML}{ff6347}  

\definecolor{cashyellow}{HTML}{F9BA2E}
\definecolor{cashblue}{HTML}{1AA6EA}
\definecolor{cpeyellow}{HTML}{fff3b0}
\definecolor{cpepink}{HTML}{ffccd5}
\definecolor{cpegreen}{HTML}{d0f4de}
\definecolor{cpelime}{HTML}{eeef20}
\definecolor{lavenderblue}{HTML}{bde0fe}

\definecolor{lightRed}{HTML}{FF9999} 
\definecolor{pastelRed}{HTML}{FF6961} 
\definecolor{softCoral}{HTML}{FF7F7F} 
\definecolor{blushPink}{HTML}{FFC0CB} 
\definecolor{lightRose}{HTML}{FFB6C1} 
\definecolor{dustyPink}{HTML}{DCAE96} 
\definecolor{powderPink}{HTML}{F8E4E3} 
\definecolor{rosyPink}{HTML}{E75480} 

\definecolor{cherryBlossom}{HTML}{FFB7C5} 
\definecolor{bubblegumPink}{HTML}{FFC1CC} 
\definecolor{carnationPink}{HTML}{FFA6C9} 
\definecolor{lightCrimson}{HTML}{F56991} 
\definecolor{pinkLace}{HTML}{FFDDF4} 
\definecolor{pastelPink}{HTML}{FFD1DC} 
\definecolor{mistyRose}{HTML}{FFE4E1} 
\definecolor{roseGold}{HTML}{B76E79} 
\definecolor{apricotBlush}{HTML}{FDD5B1} 
\definecolor{pearlPink}{HTML}{FDE0DC} 
\definecolor{strawberryCream}{HTML}{FFC4CE} 
\definecolor{watermelonPink}{HTML}{FE6F5E} 
\definecolor{flamingoPink}{HTML}{FC8EAC} 
\definecolor{softPink}{HTML}{F4C2C2} 
\definecolor{sunsetBlush}{HTML}{FFDAB3} 
\definecolor{babyPink}{HTML}{FADADD} 
\definecolor{petalPink}{HTML}{FFB3C6} 
\definecolor{cottonCandy}{HTML}{FFBCD9} 
\definecolor{peachFuzz}{HTML}{FFBE98}

\definecolor{LightThistle}{HTML}{FFDEFF}
\definecolor{Illuminate}{HTML}{F5DF4D}
\definecolor{LivingCoral}{HTML}{FA7268}

\usepackage{fancyhdr}
\usepackage[many]{tcolorbox}
\usepackage{hyperref}
\hypersetup{ 
     colorlinks=true,
     filecolor=cherryRed, 
     citecolor = MediumGreen,       
     urlcolor=coralRed, 
     } 
\usepackage{url}            
\usepackage{booktabs}       
\usepackage{amsfonts}       
\usepackage{nicefrac}       
\usepackage{microtype}      
\usepackage{lipsum}
\usepackage{fancyhdr}       
\usepackage{graphicx}       
\graphicspath{{media/}}     
\usepackage{amssymb}
\usepackage{amsmath}
\usepackage{bm}
\usepackage{xcolor}
\usepackage{setspace}
\usepackage{authblk} 
\usepackage{subcaption, multirow, lineno, bbm}
\DeclareMathAlphabet{\mathp}{U}{bbold}{m}{n}

\author[1]{Xi Chen}
\author[2]{Rahul Bhadani}
\author[1]{Larry Head}
\affil[1]{The University of Arizona, Tucson, USA}
\affil[2]{The University of Alabama in Huntsville, Huntsville, USA}

\title{Conformal Trajectory Prediction with Multi-View Data Integration in Cooperative Driving
\thanks{\textit{\underline{Citation}}: 
\textbf{Authors. Title. Pages.... DOI:000000/11111.}} 
}

\begin{document}
\maketitle

\begin{abstract}
Current research on trajectory prediction primarily relies on data collected by onboard sensors of an ego vehicle. With the rapid advancement in connected technologies, such as vehicle-to-vehicle (V2V) and vehicle-to-infrastructure (V2I) communication, valuable information from alternate views becomes accessible via wireless networks. The integration of information from alternative views has the potential to overcome the inherent limitations associated with a single viewpoint, such as occlusions and limited field of view. In this work, we introduce V2INet, a novel trajectory prediction framework designed to model multi-view data by extending existing single-view models. Unlike previous approaches where the multi-view data is manually fused or formulated as a separate training stage, our model supports end-to-end training, enhancing both flexibility and performance. Moreover, the predicted multimodal trajectories are calibrated by a post hoc conformal prediction module to get valid and efficient confidence regions. We evaluated the entire framework on the real-world V2I dataset V2X-Seq. Our results demonstrate superior performance in terms of Final Displacement Error (FDE) and Miss Rate (MR) using a single GPU. The code is publicly available at: \url{https://github.com/xichennn/V2I_trajectory_prediction}.
\end{abstract}


\section{Introduction}
\label{sec:intro}

\begin{figure}
  \begin{subfigure}{0.30\textwidth}
    \includegraphics[width=\textwidth]{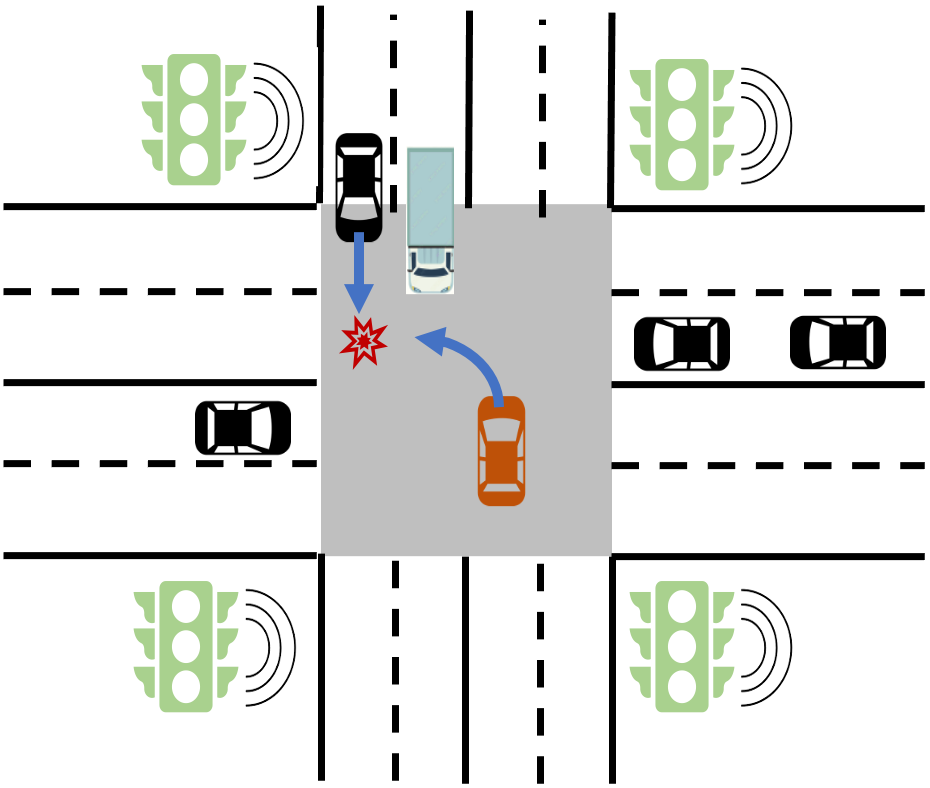} 
    \caption{potential left-turn collision}
  \end{subfigure}
  \hfill
  \begin{subfigure}{0.30\textwidth}
    \includegraphics[width=\textwidth]{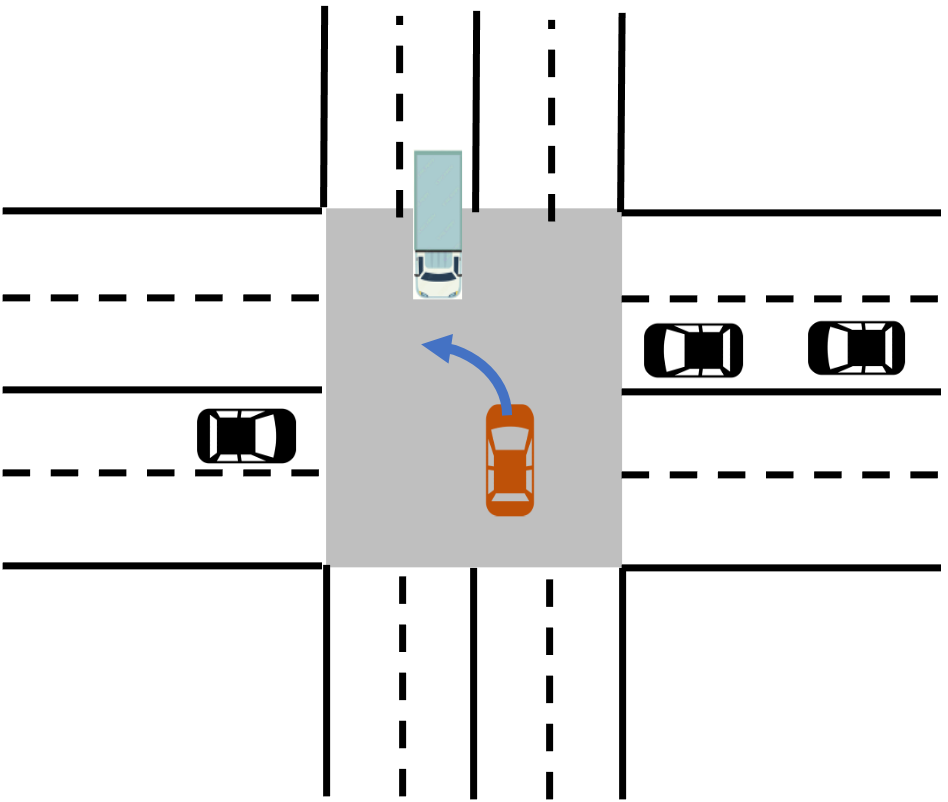}
    \caption{vehicle view}
  \end{subfigure}
  \hfill
  \begin{subfigure}{0.30\textwidth}
    \includegraphics[width=\textwidth]{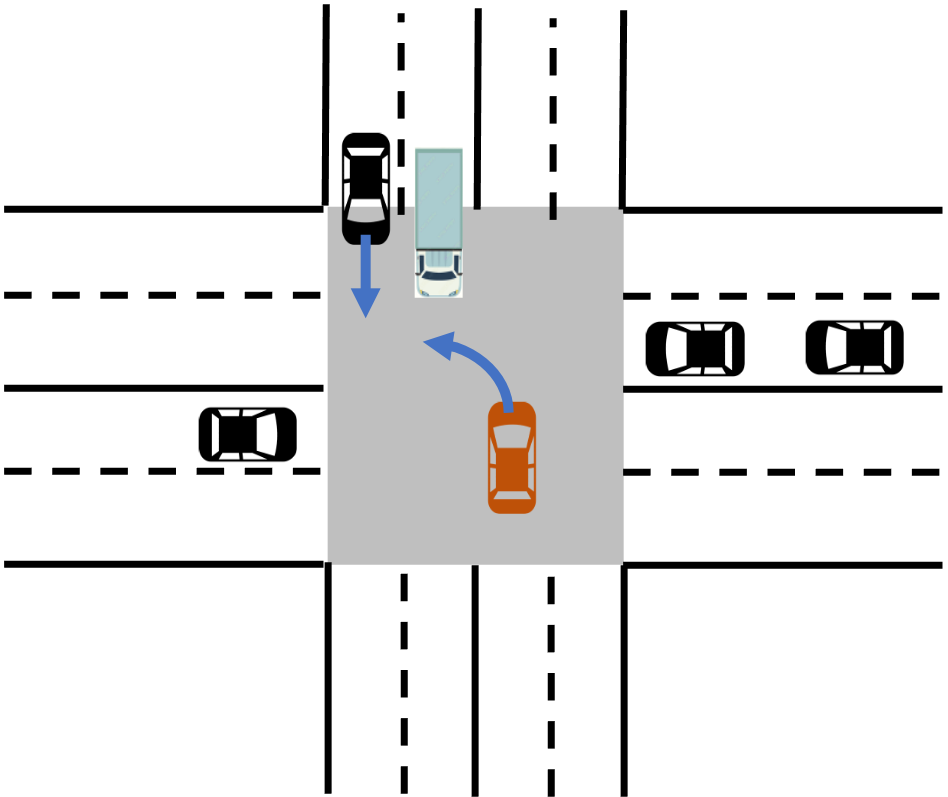}
    \caption{infrastructure view}
  \end{subfigure}
  \caption[Motivational scenarios]{Motivational scenarios. AV is in orange. (a) The AV is attempting a left turn and is at risk of a potential collision with an oncoming vehicle going straight (b) The AV’s onboard sensors have their field of view obstructed by large trucks or other vehicles, limiting their ability to detect the oncoming traffic. (c) The roadside cameras' view of the intersection. They are positioned to have an unobstructed view of the entire intersection, providing a complete picture of the traffic situation.} \label{figchap5:motivation}
\end{figure}

Trajectory prediction plays a critical role in autonomous driving. Typically, we rely on the on-board sensors of an ego vehicle to gather surrounding information necessary for performing various autonomous driving tasks. However, with the rapid advancement in connected technologies, such as vehicle-to-vehicle (V2V) and vehicle-to-infrastructure (V2I) communication, valuable information from alternate views becomes accessible via wireless networks. The integration of information from alternative views has the potential to overcome the inherent limitations associated with a single viewpoint, such as occlusions and limited field of view. We categorize the data obtained solely from ego vehicle on-board sensors or infrastructure as single-view information, whereas the accessibility to multiple viewpoints is referred to as multi-view data. Figure~\ref{figchap5:motivation} depicts an intersection scenario where the AV faces a potential left-turn collision due to its field of view being obstructed by a large truck. Roadside cameras, strategically positioned to have an unobstructed view of the entire intersection, provide a comprehensive overview of the traffic situation. This enhanced perspective allows the AV to receive critical information and effectively avoid the impending collision.
From a trajectory predictive modeling perspective, significant research efforts have been dedicated to only to use single-view datasets collected by the ego vehicle \cite{liang2020learning, chai2019multipath, zhou2022hivt, gao2020vectornet, zhao2020tnt}. These efforts typically involve modeling the temporal dependencies, agent-agent interactions, and agent-lane relations. However, the challenge arises when dealing with multi-view data, particularly in how to effectively fuse the information due to overlapping field of views.

Existing work on multi-view data fusion in the context of cooperative driving predominantly focuses on collaborative perception tasks, with limited research addressing trajectory prediction. \cite{yu2023v2x} pioneered the creation of the first V2I real-world dataset for trajectory prediction studies. They manually fused multi-view data using trajectory association and stitching techniques at each frame, followed by the application of single-view trajectory prediction models. Although this approach is intuitive, it fails to fully exploit the motion behavior captured by data from each view, resulting in suboptimal outcomes. To address this limitation, \cite{ruan2023learning} proposed a novel approach that encodes trajectory information from each view as independent graph nodes, thereby minimizing information loss. They formulated the node association problem as a graph link prediction task and introduced a cross-attention module to fuse node embeddings from associated nodes across different views. While their model can be trained end-to-end, the node association process necessitates pretraining, leading to the formulation of two optimization objectives.

We introduce a trajectory prediction framework that utilizes multi-view data without the need for explicit association between different perspectives. Rather than developing a specialized multi-view model, our approach seamlessly integrates with state-of-the-art single-view trajectory models, maximizing the utility of existing research efforts. No special training strategies are required. We can easily take advantage of the pretrained single-view trajectory models to expedite the training. Specifically, for trajectory data from each single-view, we employ established graph neural network (GNN) based models, such as LaneGCN \cite{liang2020learning}, HiVT \cite{zhou2022hivt}, to capture temporal dependencies, agent-agent interactions, and agent-lane relations. Then we utilize a cross-graph attention module to fuse the node embeddings from different views. The fused final embeddings will then go through a multimodal decoder to get future trajectory predictions. 

\begin{figure}
  \centering
  \includegraphics[width=.8\textwidth]{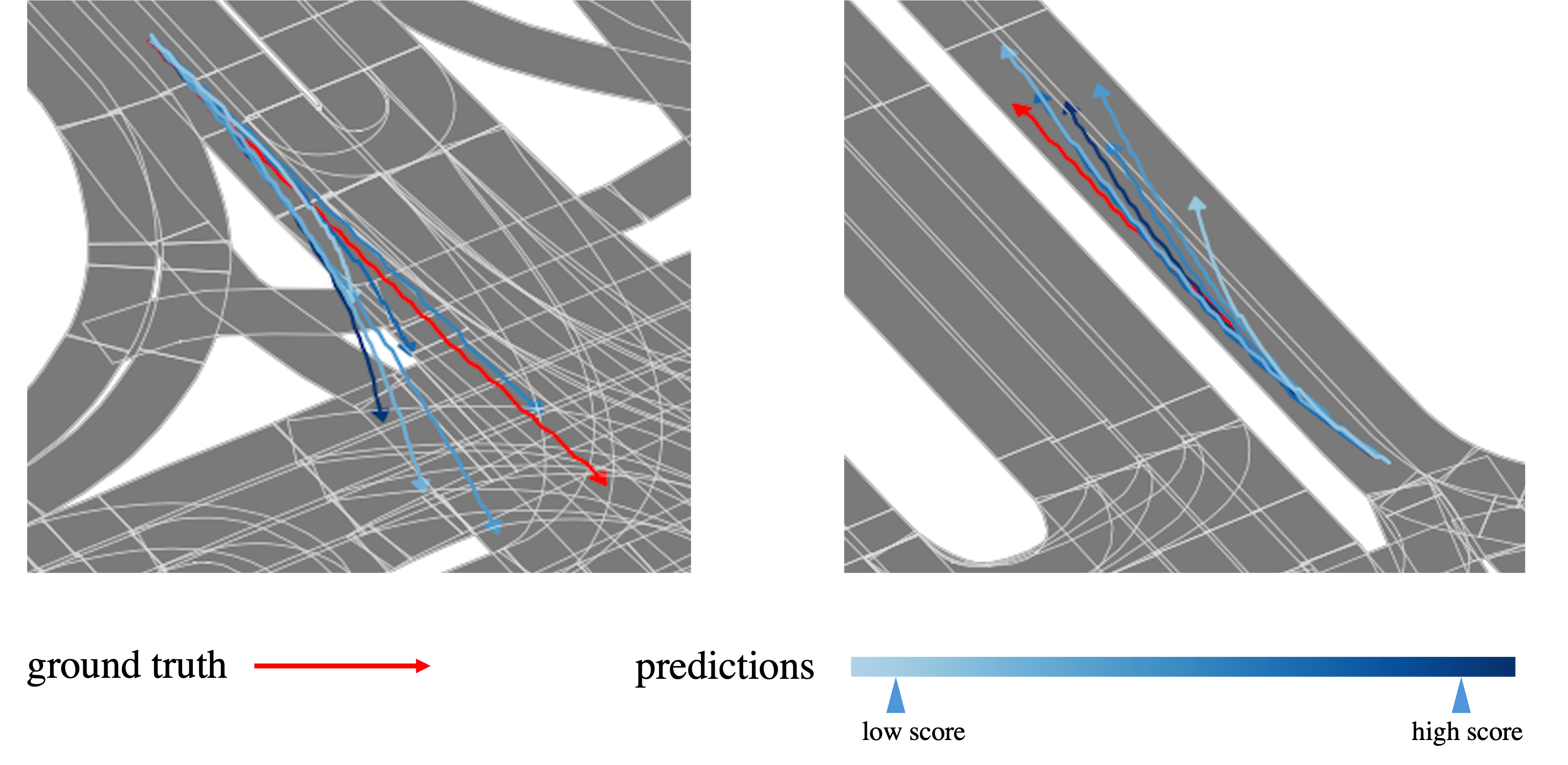}
  \caption[Inaccurate inference scores]{Inaccurate inference scores. The ground truth trajectory is represented in red, while predictions with lower scores are depicted in lighter shades of blue}\label{figchap5:infer-error}
\end{figure}

Existing works have modeled the multi-modality explicitly by introducing anchors \cite{chai2019multipath, phan2020covernet, varadarajan2022multipath++}, mixture models \cite{mercat2020multi,khandelwal2020if,buhet2020plop, cui2019multimodal, liang2020learning, zhou2022hivt}, or implicitly through latent variables such as Conditional Variational Auto-encoder (CVAE) or generative models \cite{feng2019vehicle, ivanovic2020multimodal, zhong2022stgm}. The implicit models often face the issue of mode collapse, therefore we will model the multi-modality by a mixture model built upon MLP. It is a common strategy to assign a higher score to the modality closer to the ground truth during training. The strategy, however, may encounter robustness issues during inference. Figure~\ref{figchap5:infer-error} presents two examples where the top-scored prediction is not the closest to the ground truth trajectory.

 The situation arises from the inherent randomness in the ground truth data, posing challenges for the model to learn the priorities of multiple predictions. Relying on the scores to rank the predictions can lead to hazardous results in safety-critical scenarios. Additionally, while most mixture models like the Gaussian Mixture Model (GMM) provide both point estimation and variance estimation, the reliability of these uncertainty quantifications (UQ) remains questionable. Unfortunately, this aspect is often overlooked in the model evaluation process. However, UQ can offer valuable insights for downstream decision-making processes. To address these challenges, we introduce conformal prediction (CP) \cite{shafer2008tutorial, papadopoulos2007conformal}, as a post-hoc module to our framework. CP creates statistically rigorous uncertainty intervals for the predictions that are guaranteed to contain the ground truth with a user-specified probability. 

To summarize, our contributions in this work lie in twofolds:
1) We introduce V2INet, a novel trajectory prediction framework designed to model multi-view data by extending existing single-view models. Unlike previous work, our model supports end-to-end training, enhancing both flexibility and performance.
2) We introduce conformal prediction as a post-hoc module to calibrate the prediction results. This results in statistically rigorous uncertainty intervals, significantly enhancing the reliability of the predictions. These calibrated predictions are particularly beneficial for downstream decision-making tasks, such as motion planning.

\section{Related Work}
\subsection{Cooperative Driving Dataset}
Collaborative perception has been the most extensively studied area in collaborative driving \cite{kim2015impact, cui2022coopernaut, cui2022cooperative, ren2022collaborative}. It uses the recent developments in wireless communication technologies, such as V2V and V2I, to share information, which enables perception beyond line-of-sight and field-of-view hence overcoming common perceptual shortcomings with individual perception, such as blindspot, occlusions, and long-range issues \cite{wang2020v2vnet,furst2020lrpd}. Many datasets have been collected in both simulation environment \cite{arnold2020cooperative, xu2022opv2v, li2022v2x} and real-world \cite{yu2022dair,xu2023v2v4real} to facilitate the study. A comprehensive comparison can be found in Table~\ref{tabchap5:dataset}. While a majority of them focused on the upstream tasks such as detection, tracking, and segmentation, to enable well-informed decision-making for autonomous vehicles, however, it's critical to also incorporate V2X data for predicting the behavior of surrounding traffic participants. V2X-Seq/Forecasting \cite{yu2023v2x} and V2X-Traj \cite{ruan2023learning} present two real-world datasets specifically designed for cooperative trajectory prediction tasks. The experiments in V2X-Seq/Forecasting demonstrate that leveraging the infrastructure-side trajectories can enhance the trajectory prediction performance. V2X-Traj aims for more general V2X scenarios, extending to V2V cooperation.

\begin{table}[h!]
    \centering
    \caption[Cooperative Driving Datasets Comparison]{Comparison of Cooperative Driving Datasets. The compared tasks include detection (Det.), tracking (Track), segmentation (Seg.) and trajectory prediction (Pred). Only the tasks performed in the original paper are reported. "-" means the number was not reported."*" means expected.}
    \label{tabchap5:dataset}
    \resizebox{\textwidth}{!}{%
    \begin{tabular}{cccccccccccc}
    \hline 
    \hline
     \multirow{2}{*}{Dataset} & \multirow{2}{*}{Year} & \multirow{2}{*}{Source} & \multirow{2}{*}{Scenario}   
     & \multicolumn{4}{c}{Tasks} & \multirow{2}{*}{Maps}  & \multirow{2}{*}{Total Time} & \multirow{2}{*}{Traffic Signal} & \multirow{2}{*}{Scenes} \\
     \cline{5-8}
     \multicolumn{4}{c}{} & Det. & Track. & Seg. & Pred. & \multicolumn{4}{c}{} \\
     \hline
     Cooper(inf)&2019&CARLA&V2I& \checkmark & x & x & x & x &-& x & \textless 100 \\
     DAIR-V2X-C & 2021 & Real-world & V2I & \checkmark & x & x & x & \checkmark & 0.5h & x & 100\\
     OPV2V & 2021 & CARLA+OpenCDA & V2V & \checkmark & x & x & x & x & 0.2h & x & 73\\
     V2X-Sim & 2022 & CARLA+SUMO & V2V\&I & \checkmark & \checkmark & \checkmark & x & x & 0.3h & x & 100\\
     V2V4Real & 2023 & Real-world & V2V & \checkmark & \checkmark & x & x & \checkmark & 19h & x & 67 \\
     V2X-Seq/Perception & 2023 & Real-world & V2I & x & \checkmark & x & x & \checkmark & 0.43h & x & 95\\
     V2X-Seq/Forecasting & 2023 & Real-world & V2I & x & x & x & \checkmark & \checkmark & 583h & \checkmark & 210000\\
     V2X-Traj & 2024* & Real-world & V2V\&I & x & x & x & \checkmark & \checkmark &-& \checkmark & 6160\\
    \hline
    \end{tabular}%
    }
\end{table}

\subsection{Cooperative Information Association and Fusion}

The main challenge in cooperative trajectory prediction lies in the multi-view data source fusion. In cooperative scenarios, vehicles gather safety-related data using sensors like radar, lidar, cameras, and GPS. This data is standardized into Basic Safety Message (BSM) format, ensuring compatibility across vehicles and infrastructure \cite{kamrani2018extracting}. BSM messages, containing crucial information such as position and speed, are broadcasted periodically. Upon receiving BSM messages, vehicles combine this data with their own sensor data to enhance accuracy. Advanced algorithms have been developed to associate and fuse the multi-source data. Given the temporal and spatial dimensions of the collected trajectories, there has been research focusing on the communication delays alignment \cite{lei2022latency,wang2020v2vnet,xu2022v2x} and pose errors alignment \cite{yuan2022leveraging,lu2023robust,besl1992method,wang2020v2vnet}. Based on the aligned data, \cite{yu2023v2x} utilized CBMOT \cite{benbarka2021score}, a multi-object tracking method, to fuse the infrastructure and ego-vehicle trajectories at each single frame and then trained the network taking in the fused dataset. While straightforward, this method fails to capture the motion behavior provided by the infrastructure across all time steps, leading to suboptimal results. In contrast, \cite{ruan2023learning} encoded trajectory information from each view as independent graph nodes, formulating the association process as a graph linking problem. While effective, this approach necessitates separate training procedures. Building upon their work, we propose the utilization of a cross-graph attention mechanism to fuse multi-view information, eliminating the need for additional training processes.
\subsection{Uncertainty Quantification}
In the trajectory prediction task, numerous sources of uncertainty exist, including inherent multi-modality, partial observability, short time scales, data limitations, intention type imbalances, and domain gaps. Most existing trajectory prediction models address uncertainty by maximizing the likelihood of an assumed distribution, such as Gaussian or Laplace \cite{varadarajan2022multipath++, zhou2022hivt, liang2020learning, khandelwal2020if}. However, trajectories with the largest likelihood are often nonsensical \cite{zecchin2024forking, holtzman2019curious}. Alternatively, some research focuses on approximating Bayesian inference for deep learning models using techniques like Monte Carlo dropout \cite{gal2016dropout}, which involves performing stochastic forward passes through the network and averaging the results. Among these approaches, \cite{tang2023collaborative} stands out as the only work that incorporates collaborative uncertainty among agents into the modeling process to guide the ranking of multimodal trajectories by uncertainty, albeit requiring special training strategies. However, none of the predicted uncertainties from these methods offer finite sample coverage guarantees, which is suboptimal for safety-critical applications such as vehicle trajectory prediction.

Conformal prediction (CP) \cite{papadopoulos2007conformal, angelopoulos2023conformal} has emerged as a widely adopted uncertainty quantification method, owing to its simplicity, generality, theoretical rigor, and low computational overhead. Notably, CP is agnostic to the underlying model and data distribution, making it highly versatile. It seamlessly integrates with any pre-trained model to deliver statistically valid prediction regions. Of particular relevance to our multimodal trajectory prediction task is recent progress in generalizing CP to time-series forecasting. For instance, \cite{sun2022copula} introduced the Copula conformal prediction algorithm for multivariate, multi-step time series forecasting, applicable to any multivariate multi-step forecaster. Additionally, \cite{tumu2023multi} focused on generating non-conformity score functions that yield multimodal prediction regions with minimal volume. Moreover, \cite{khurjekar2024multi} employed CP to generate statistical uncertainty intervals from Gaussian mixture model outputs, obtaining separate prediction intervals corresponding to each GMM component prediction. Furthermore, \cite{huang2024uncertainty} proved the validity of CP on graph data. Inspired by their work, we explore the potential of applying CP to the multimodal trajectory prediction comparing different CP methods.

\section{Problem Formulation}
\label{chap5sec:problemformulation}
At the scenario level, We have trajectory data $\mathcal{T}$ from both the vehicle and infrastructure viewpoints, denoted as $\mathcal{T^V}$ and $\mathcal{T^I}$, respectively. While $\mathcal{T^V}$ and $\mathcal{T^I}$ share overlapping information where their fields of view intersect, $\mathcal{T^I}$ also provides complementary information, being free from occlusions. Our modeling objective is to utilize the information from $\mathcal{T^I}$ to improve the accuracy of trajectory prediction based solely on $\mathcal{T^V}$. 

The trajectory prediction task involves leveraging historical trajectories $\mathcal{T}^\mathcal{V}_h \in \mathbb{R}^{N^\mathcal{V} \times T_h \times a_h}$ and $\mathcal{T}^\mathcal{I}_h \in \mathbb{R}^{N^\mathcal{I} \times T_h \times a_h}$, alongside contextual information, typically HD maps denoted as $\mathcal{M}$, to forecast future trajectories $\mathcal{T}^\mathcal{V}_f \in \mathbb{R}^{N^\mathcal{V} \times T_f \times a_f}$. Here, $N^\mathcal{V}$ and $N^\mathcal{I}$ represent the number of observed actors from the vehicle and infrastructure perspectives, respectively. $T_h$ denotes the historical time horizon, and $T_f$ is the prediction horizon. $a_h$ and $a_f$ represent the number of node features which we consider the vehicle center location defined by its $x-$ and $y-$coordinates. Notably, the trajectory data from both views are defined within the same coordinate system. For the HD map, we opt for a vectorized representation due to its lightweight nature and efficiency \cite{gao2020vectornet}. This vector map is depicted by lane centerlines, which are composed of lane segments. We denote it as $\mathcal{M} \in \mathbb{R}^{N^l \times a_l}$, where $N^l$ is the number of lane segments and $a_l$ is the number of lane attribute.

We formulate the overall probabilistic distribution as $\mathp{P}(\mathcal{T}^\mathcal{V}_f | \mathcal{T}^\mathcal{V}_h, \mathcal{T}^\mathcal{I}_h, \mathcal{M}^\mathcal{V}, \mathcal{M}^\mathcal{I})$. Driven by the critical safety demands inherent in trajectory prediction, we aim to incorporate uncertainty quantification to preempt any potentially consequential model failures. Let $\mathcal{D}$ be the set of scenarios of the form $(\mathcal{T}^\mathcal{V}_h, \mathcal{T}^\mathcal{I}_h, \mathcal{M}^\mathcal{V}, \mathcal{M}^\mathcal{I}, \mathcal{T}^\mathcal{V}_f)$. We split the dataset into training $\mathcal{D}_{train}$, validation $\mathcal{D}_{val}$, calibration $\mathcal{D}_{cal}$ and test $\mathcal{D}_{test}$. One black-box deep learning model is trained on $\mathcal{D}_{train}$ and evaluated on $\mathcal{D}_{val}$. We achieve the uncertainty quantification by conformal prediction in a post-hoc way on $\mathcal{D}_{cal}$.  

At the agent level, we denote the features and labels of agent $i$ from the vehicle view as $X^\mathcal{V} = \mathcal{T}^\mathcal{V}_{h,i}$ and $Y^\mathcal{V} = \mathcal{T}^\mathcal{V}_{f,i}$ for brevity. In real-world scenarios, future trajectories may exhibit multimodal behavior, often approximated by mixture models, resulting in $\hat{Y}^\mathcal{V} \in \mathbb{R}^{K \times T_f \times a_f}$, where $K$ represents the number of mixtures or modes. Given a new agent sample $X^\mathcal{V}_{test}$ from $\mathcal{D}_{test}$, we seek to construct the prediction intervals $\mathcal{C}(X^\mathcal{V}_{test}) \in \mathbb{R}^{K \times T_f \times a_f \times 2}$ such that it covers the ground truth label $Y^\mathcal{V}_{test}$ under a predefined coverage rate leveraging conformal prediction.

CP proceeds in three steps. First, we define a nonconformity score $A: \mathcal{X} \times \mathcal{Y} \in \mathbb{R}^{K \times T_f \times a_f}$ to quantify how well $Y$ conforms to the prediction at $X$. Typically, we choose a metric of disagreement between the prediction and the ground truth as the non-conformity score, such as the Euclidean distance. Second, given the predefined miscoverage rate $\alpha$, we compute the $1-\alpha$ quantile of the non-conformity scores on the calibration set $\{A(X_1, Y_1),\cdots,A(X_n, Y_n)\}$, where $n$ is the number of agents. The resulting quantile is denoted as $\hat{H} \in \mathbb{R}^{T_f \times a_f}$. Last, when presented with a new test agent $X^\mathcal{V}_{test}$, CP constructs the prediction interval $\mathcal{C}(X^\mathcal{V}_{test})=\{Y^\mathcal{V} \in \mathcal{Y}: A(X^\mathcal{V}_{test}, Y^\mathcal{V}) \leq \hat{H}\}$.

\section{Methodology}
\subsection{Overview}
\begin{figure}[!ht]
  \centering
  \includegraphics[width=\textwidth]{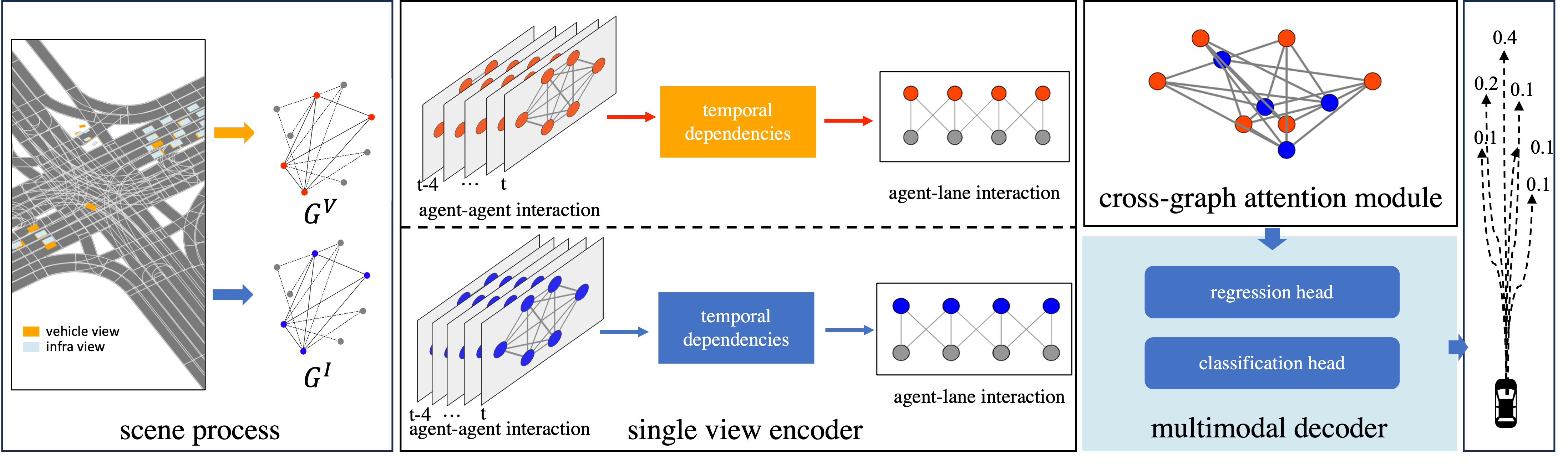}
  \caption[Proposed multi-view model architecture]{Proposed model architecture. Data collected from the vehicle view are represented in red, while data from the infrastructure view are depicted in blue. The model takes as input the graph data constructed from both views. We apply single-view encoders to encode information from each view, followed by the fusion of the two embeddings through a cross-graph attention module. The final embedding passes through a multi-modal decoder, providing multimodal predictions for all the agents of interest.}\label{figchap5:arch}
\end{figure}

Our method V2INet consists of two key components, predictive modeling and post-hoc conformal prediction. An overview of our proposed model is illustrated in Figure~\ref{figchap5:arch}. 
We first represent the scenario data collected from both views as graphs. A single view encoder is then applied separately to each graph, encoding various information such as agent-agent interactions, temporal dependencies, and agent-lane information. Subsequently, The vehicle-view embedding is fused with the infrastructure-view embedding through a cross-graph attention module. Finally, the updated vehicle-view embedding pass through a multi-modal decoder, providing multimodal predictions for all the agents of interest. The post-hoc conformal prediction is then applied at the agent level to construct valid prediction intervals given a predefined coverage rate.

\subsection{Scene Representation}
We adopt an ego-centric coordinate system utilizing vectorized representation as first introduced in \cite{gao2020vectornet}. First, data from both views are transformed such that the ego vehicle is centered at the origin, with its heading aligned along the positive $x$-axis. Each trajectory is then characterized as a sequence of displacements  $\{{\Delta \mathbf{x}^{t}\}}^0_{t=-(T_h-1)}$, where $\Delta \mathbf{x}^{t} \in \mathbb{R}^2$ is the 2D displacement from time step $t-1$ to $t$. Similarly, each lane segment is represented as $\Delta \mathbf{x}^l \in \mathbb{R}^2$, which captures the 2D displacement from the starting coordinate to the end coordinate of lane segment $l$. We then construct scenario graphs consisting of actor nodes and lane nodes for both views separately. The edge attribute is the absolute relative position between two nodes.

\subsection{Single View Encoder}
To encode the spatiotemporal information, including agent-agent interactions, temporal dependencies, and context information captured by agent-lane relations for data from each view, existing graph-based models offer effective solutions. Models like LaneGCN \cite{liang2020learning}, HiVT \cite{zhou2022hivt}, and VectorNet \cite{gao2020vectornet} incorporate these components and can be readily employed. This encoding step can be performed efficiently at the edge device, such as the vehicle's on-board computers and the roadside unit (RSU) for infrastructure data, before broadcasting, thereby minimizing computational overhead.

We exemplify here with HiVT \cite{zhou2022hivt} where only attention mechanism is employed, with the adoption of a rotation-invariant representation. To attend to the local information, at each timestamp, the surrounding actors' information is aggregated and the embeddings at each time stamp subsequently go through a transformer to capture the temporal dependencies. After obtaining the spatio-temporal embeddings, the surrounding lane information is aggregated for each actor. The update actor node embeddings is then passed to a global interaction module which aggregates the local context of different actors and updates each actor’s representation to capture long-range dependencies and scene-level dynamics. We denote the final embedding for actor node $i$ from the vehicle view as $\mathbf{h}_i$, and actor node $j$ from the infrastructure view as $\mathbf{h}_j$. Both $\mathbf{h}_i$ and $\mathbf{h}_j$ are in $\mathbb{R}^{d_h}$, where $d_h$ is the embedding dimension.

\subsection{Cross-graph Attention Module}
To effectively utilize information from the infrastructure side, we employ a cross-graph attention module that aggregates information captured by the infrastructure view encoder. Specifically, the embedding from the vehicle view $\mathbf{h}_i$ is transformed into the query vector, while the embedding from the infrastructure view $\mathbf{h}_j$ alongside the relative position at the last observed time step $\mathbf{x}_i^{t=0} - \mathbf{x}_j^{t=0}$, are utilized to compute the key and value vectors. We denote $\mathbf{h}_{ij}=(\mathbf{h}_j, \mathbf{x}_i^{t=0} - \mathbf{x}_j^{t=0})$ representing the concatenation of node and edge attribute from infrastructure node $j$ to vehicle node $i$:
\begin{equation}
    \mathbf{q}_i=\mathbf{W}^{Q_f}\mathbf{h}_i,  \quad \mathbf{k}_{ij}=\mathbf{W}^{K_f}\mathbf{h}_{ij}, \quad \mathbf{v}_{ij}=\mathbf{W}^{V_f}\mathbf{h}_{ij}
\end{equation}
where $\mathbf{W}^{Q_f}$, $\mathbf{W}^{K_f}$, $\mathbf{W}^{V_f} \in R^{d_k \times d_h}$ are learnable matrices for linear projection and $d_k$ is the transformed dimension. The resulting query, key and value vectors are then taken as input to the scaled dot-product attention block:
\begin{linenomath}
    \begin{flalign} 
        &\mathbf{\alpha}_{ij} = \text{softmax}\left(\frac{{\mathbf{q}_i}^T}{\sqrt{d_k}} \cdot \Big[\{\mathbf{k}_{ij}\}_{j \in \mathcal{N}_i}\Big]\right),\\
        &\mathbf{m}_i = \sum_{j \in \mathcal{N}_i}\mathbf{\alpha}_{ij}\mathbf{v}_{ij},\\
        &\mathbf{g}_i = \text{sigmoid}(\mathbf{W}^{\text{gate}}[\mathbf{h}_i,\mathbf{m}_{i}]),\\
        &\tilde{\mathbf{h}}_i = \mathbf{g}_i \odot \mathbf{W}^{\text{self}}\mathbf{h}_i+(1-\mathbf{g}_i)\odot\mathbf{m}_i
    \end{flalign}
\end{linenomath}
where $\mathcal{N}_i$ is the set of agent $i$'s neighbors, $\mathbf{W}^{\text{gate}}$ and $\mathbf{W}^{\text{self}}$ are learnable matrices and $\odot$ denotes element-wise product. We followed the structure in HiVT to fuse the infrastructure information with a gating function. The attention block supports multiple heads. Finally, we apply another MLP block to obtain the final fused embedding $\tilde{\mathbf{h}}_i \in \mathbb{R}^{d_h}$ for agent~$i$ from the vehicle view. 
\subsection{Mixture Model Based Decoder and Learning}
There are two widely used mixture models for describing the multimodal trajectories, Gaussian and Laplacian. Previous methods \cite{liang2020learning, gao2020vectornet, kosaraju2019social} have found that the $\ell_1$-based loss function derived from the Laplace distribution usually leads to superior prediction performances, as it is more robust to outliers. Hence, we will parameterize the future trajectories following Laplace distribution. For each agent, the decoder receives the final embedding as inputs and outputs $K$ possible future trajectories and the mixing coefficient of the mixture model for each agent. The decoder are consisted of three MLPs, one for predicting the future locations $\mathbf{\mu}^t_{i,k} \in \mathbb{R}^2$ for agent $i$ and its mode $k$ at each time step $t$, one for predicting the associated uncertainty $\mathbf{b}^t_{i,k} \in \mathbb{R}^2$ assuming independence of the $x-$ and $y-$coordinates, the last one followed by a softmax is for producing the scores for each mode. 

To ensure the prediction diversity \cite{gupta2018social, thiede2019analyzing}, instead of optimizing all the predicted trajectories, only the mode $\tilde{k}$ closest to the ground truth is optimized. The closeness here is defined as the average Euclidean distance between ground-truth locations and predicted locations across all future time steps. The Loss includes both regression loss and classification loss
\begin{linenomath}
    \begin{equation}
      J = J_{reg} + \varepsilon J_{cls} 
    \end{equation}
\end{linenomath}
Here, $\varepsilon$ is the weight of the classification loss. We employ the negative log-likelihood as the regression loss:
\begin{linenomath}
    \begin{equation}
      J_{reg} = -\frac{1}{nT_f}\sum^n_{i=1} \sum^{T_f}_{t=1} \text{log}\mathbf{P}(\mathbf{y}_i^t - \hat{\mathbf{y}}_i^t | {\hat{\mathbf{\mu}}}^{t,\tilde{k}}_i,\hat{\mathbf{b}}^{t,\tilde{k}}_i)
    \end{equation}
\end{linenomath}

where $\mathbf{P}(|)$ is the probability density function of Laplace distribution and $\hat{\mathbf{\mu}}_i^{t,\tilde{k}}$, $\hat{\mathbf{b}}_i^{t,\tilde{k}}$ are the mean and uncertainty estimates of the best mode $\tilde{k}$.

For $J_{cls}$, cross-entropy loss is applied:
\begin{linenomath}
    \begin{equation}
      J_{cls} = -\frac{1}{n}\sum^n_{i=1} \sum^{K}_{k=1} \mathbbm{1}_k \text{log}{\pi_k}
    \end{equation}
\end{linenomath}
where $\mathbbm{1}_k$ is a binary indicator if $k$ is the best mode, $\pi_k$ is the mixing coefficient of mode $k$. The model is trained on $\mathcal{D}_{train}$ and evaluated on $\mathcal{D}_{val}$.

\subsection{Post-hoc uncertainty quantification module}
Uncertainty quantification methods are evaluated on two key properties: validity and efficiency. Validity is established when the predicted confidence level exceeds or equals the probability of events falling within the predicted range, while efficiency refers to minimizing the size of the confidence region. We utilize conformal prediction to obtain both valid and efficient prediction intervals. The standard conformal prediction methods typically operate with scalar point estimates for regression problems. However, since our output consists of a multimodal multivariate time-series, we need to make certain adaptations to accommodate this complexity. Following the steps outlined in Section~\ref{chap5sec:problemformulation}, we proceed with non-conformity score function definition, quantile computation and prediction interval construction.
\subsubsection{Non-conformity Score Functions}
As emphasized in \cite{angelopoulos2021gentle}, the usefulness of the prediction sets is primarily determined by the score function, we adopt three score functions $A$ and compare their performance.

\textbf{Z-score}. As the decoder returns both mean and variance predictions, we define the Z-score function as:
\begin{equation}
  Z = \frac{|Y -\hat{Y}|}{\hat{B}}
\end{equation}

where $Z \in \mathbb{R}^{K \times T_f \times 2}$ and $\hat{B} = \{\{\hat{b^{t,k}}\}_{t=0}^{T_f}\}_{k=0}^K$. Here, we compute scores separately for $x$- and $y$-coordinates, reflecting the observation that motion uncertainty varies significantly across different dimensions.

\textbf{L2-norm}. Next, we consider the Euclidean distance, the most commonly used metric in regression problems:
\begin{equation}
  L_2 = ||Y -\hat{Y}||_2
\end{equation}
where $L_2 \in \mathbb{R}^{K \times T_f}$.

\textbf{L1-norm}. Recognizing that the L2-norm disregards dimension differences, we consider L1-norm:
\begin{equation}
  L_1 = ||Y -\hat{Y}||_1
\end{equation}
where $L_1 \in \mathbb{R}^{K \times T_f \times 2}$.

\subsubsection{Quantile Computation}
Given a predefined miscoverage rate $\alpha \in [0,1]$, the $1-\alpha$ quantile of the non-conformity scores is calculated on the calibration set $\mathcal{D}_{cal}$. Conventionally, the quantile is determined as follows: $\hat{H}=\text{quantile}(\{A(X_1, Y_1),\cdots,A(X_n, Y_n)\}, (1-\alpha)(1+\frac{1}{n}))$ where $A(X_i, Y_i)$ is a scalar, and $n$ denotes the total number of agents in $\mathcal{D}_{cal}$. However, in our case, we deal with multimodal time-series scores, which are multivariate when using Z-score and L1-norm, hence necessary adaptations are needed. For brevity, let $\Gamma_i = A(X_i, Y_i)$, with $\Gamma_i \in \mathbb{R}^{K \times T_f \times 2}$ for Z-score and L1-norm, and $\mathbb{R}^{K \times T_f}$ for L2-norm.

To enhance the efficiency of our prediction intervals, we focus on computing the quantile using only the mode $\tilde{k}$ that exhibits the smallest average Euclidean distance to the ground truth trajectory. This reduces the computation to multivariate time-series. Within this framework, we investigate two established methods: CF-RNN \cite{stankeviciute2021conformal} and CopulaCPTS \cite{sun2022copula}.

\textbf{CF-RNN}. Since the time-series predictions are obtained from the same embedding, this work proposes the application of Bonferroni correction to the calibration scores to maintain the desired miscoverage rate $\alpha$. Specifically, the original $\alpha$ is divided by $T_f$, yielding $\hat{H}=\text{quantile}(\{\Gamma_i\}_{i=0}^n, (1-\frac{\alpha}{T_f})(1+\frac{1}{n}))$ for single-variate case.

\textbf{CopulaCPTS}. As implied by its name, this method models the joint probability of uncertainty for multiple predicted time steps using a copula. The calibration set $\mathcal{D}_{cal}$ is split into two subsets: $\mathcal{D}_{cal-1}$, which estimates a Cumulative Distribution Function (CDF) for the nonconformity score of each time step, and $\mathcal{D}_{cal-2}$, utilized to calibrate the copula. The copula function captures the dependency between time steps and can enhance the efficiency of prediction intervals.

For multivariate scores generated by the L1-norm and Z-score functions, we adopt the Bonferroni correction for each dimension, as inspired by CF-RNN. Specifically, we use $\alpha/2$ as the miscoverage rate for both $x-$ and $y-$coordinates.

\subsubsection{Prediction Interval Construction}
We utilize the obtained quantile $\hat{H}$ to form the prediction intervals for new examples in $\mathcal{D}_{test}$:
\begin{equation}
    \mathcal{C}(X^\mathcal{V}_{test})=\{Y^\mathcal{V} \in \mathcal{Y}: A(X^\mathcal{V}_{test}, Y^\mathcal{V}) \leq \hat{H}\}
\end{equation}
Specifically, for $\hat{H}$ obtained from the Z-score, we have:
\begin{equation}
    \mathcal{C}(X^\mathcal{V}_{test})=\Bigl[\hat{Y}^\mathcal{V}_{test} - \hat{B}\hat{H}, \hat{Y}^\mathcal{V}_{test} + \hat{B}\hat{H}\Bigr]
\end{equation}
and for $\hat{H}$ obtained from the L2-norm and L1-norm, we have 
\begin{equation}
    \mathcal{C}(X^\mathcal{V}_{test})=\Bigl[\hat{Y}^\mathcal{V}_{test} - \hat{H}, \hat{Y}^\mathcal{V}_{test} + \hat{H}\Bigr]
\end{equation}
\section{Experiments}
\subsection{Experimental Setup}
In this section, We introduce the specifics of the dataset, the evaluation metrics and the implementation details including hardware, hyperparameters, etc.
\subsubsection{Dataset}
We evaluate the proposed model on the publicly available large-scale and real-world V2I dataset V2X-Seq \cite{yu2023v2x}, which provides the trajectories of agents from both vehicle and infrastructure sides, along with vector map data. V2X-Seq consists of 51,146 V2I scenarios, where each trajectory is 10 seconds long with a sampling rate of 10 Hz. The task involves predicting the motion of agents in the next 5 seconds, given initial 5-second observations from both infrastructure and vehicle sides. The dataset has been split into train and validation. Our trained model is evaluated on the validation set, allowing for comparison with existing models. For post-hoc conformal prediction, we divide the validation set into calibration set and test set at a 4:1 ratio. For a discussion on the calibration data size, please refer to \cite{angelopoulos2021gentle}.

\subsubsection{Evaluation Metrics} 
\textbf{Model metrics}. For model evaluation, the standard metrics in motion predictions are adopted, including minimum Average Displacement Error (minADE), minimum Final Displacement Error (minFDE), and Miss Rate (MR), where errors between the best predicted trajectory among the K=6 modes and the ground truth trajectory are calculated. The best here refers to the trajectory that has the minimum endpoint error. The ADE metric calculates the L2 distance across all future time steps and averages over all scored vehicles within a scenario, while FDE measures the L2 distance only at the final future time step and summarizes across all scored vehicles. MR refers to the ratio of actors in a scenario where FDE are above 2 meters. 

\textbf{Conformal prediction metrics}. We assess validity and efficiency for each method. Validity is evaluated by reporting independent and joint coverage on the test set, aiming for coverage levels close to the desired confidence level $1-\alpha$. The independent coverage for each agent is calculated as:
\begin{equation}
    \text{ind. coverage}_{1-\alpha} = \frac{1}{T_f}\sum_{{X,Y \in \mathcal{D}_{test}}}\mathbbm{1}(Y \in \mathcal{C}(X))
\end{equation}
We identify the maximum independent coverage among the $K$ modes for each agent. For all metrics, the final reported value is the average among all agents in $\mathcal{D}_{test}$.

For efficiency, we calculate the average size of the predicted 2D area across all time steps for the mode $\tilde{k}$ with the maximum independent coverage:
\begin{equation}
    \text{size} = \frac{1}{T_f}||\mathcal{C}(X)_{\tilde{k}}||
\end{equation}
The 2D area is defined as ellipsis for Z-score and L1-norm, and circle for L2-norm.

The joint coverage is defined as if there exists a mode $k$ such that the truth values across all time steps fall into the confidence region:
\begin{equation}
    \text{joint coverage}_{1-\alpha} = \mathbbm{1}(\exists{k}\in K: \forall{t}\in T_f: Y_{k,t} \in \mathcal{C}(X))
\end{equation}

\subsubsection{Implementation Details}
\textbf{Model}. The model was trained for 64 epochs on an Nvidia V100S GPU with 32GB memory using AdamW optimizer \cite{loshchilov2017decoupled}. Hyperparameters including batch size, initial learning rate, weight decay and dropout rate are 32, 1e-3, 1e-4 and 0.1, respectively. The learning rate is decayed using the cosine annealing scheduler \cite{loshchilov2016sgdr}. Our model employs the original setting of HiVT with 64 hidden units for the single view encoder and 1 layer of cross-graph attention module with 8 heads. The radius of all local regions is 50 meters. The number of prediction modes is set to 6.

\textbf{Conformal prediction}. With the pretrained model and datasets $\mathcal{D}_{cal}$ and $\mathcal{D}_{test}$, we execute all conformal prediction methods on the CPU. We evaluate the methods on the three defined score functions at three different $\alpha$ levels: 0.2, 0.1, 0.05. The optimization step in CopulaCPTS remains consistent with the original work.

\subsection{Results Analysis}
In this section, we first examine the model's prediction performance, assessing it from both quantitative and qualitative perspectives. Next, we showcase the effectiveness of the post-hoc uncertainty quantification method.
\subsubsection{Model Performance}
\textbf{Comparison with state-of-the-art}. We benchmark our proposed model against state-of-the-art models using the V2X-Seq dataset, as detailed in \cite{yu2023v2x}. Results are summarized in Table~\ref{tabchap5:quanres}. Both TNT \cite{zhao2020tnt} and HiVT \cite{zhou2022hivt} are single-view models. \cite{yu2023v2x} evaluated them under two settings: Ego, where only the vehicle-view data is utilized, and PP-VIC, which employs a two-stage method with both vehicle view and infrastructure data. Specifically, in PP-VIC, data from both views are fused offline with some tracking method and then the stitched trajectories were fed into a single-view model. We retrained the HiVT model under both Ego and the PP-VIC setting with 64 epochs and present our results in Table~\ref{tabchap5:quanres}. Comparing the Ego and PP-VIC results, it's uncovered by \cite{yu2023v2x} that integrating information from the infrastructure side can enhance prediction accuracy. V2X-Graph represents the current state-of-the-art model, employing a graph link prediction module to associate two-view data, followed by fusing the embeddings of the associated nodes using attention mechanism. However, this node association module requires pre-training, leading to a two-stage training process.  Our method demonstrates the best performance in terms of minFDE and MR. Without the explicit node association from both views, our attention based fusion module can attend to the most relevant nodes from both views through learning, which significantly simplifies the modeling framework and facilitates ease of training, all while achieving better results.  
\begin{table}[htbp]
\caption[Quantitative results on V2X-Seq]{Quantitative results on V2X-Seq. TNT and V2X-Graph results are reported in \cite{yu2023v2x}.\label{tabchap5:quanres}}
\begin{center}
\resizebox{0.5\textwidth}{!}{%
\begin{tabular}{cc|ccc}
\hline 
Method&Cooperation&minADE&minFDE&MR \\
\hline
TNT&Ego&8.45&17.93&0.77\\
TNT&PP-VIC&7.38&15.27&0.72\\
\hline
HiVT&Ego&1.34&2.16&0.31\\
HiVT&PP-VIC&1.28&2.11&0.31\\
\hline
\multicolumn{2}{c|}{V2X-Graph}&\textbf{1.17}&2.03&0.29\\
\hline
\multicolumn{2}{c|}{V2INet (Ours)}&1.19&\textbf{1.98}&\textbf{0.27}\\
\hline
\end{tabular}%
}

\end{center}
\end{table}

\begin{figure}[!ht]
  \centering  \includegraphics[width=\textwidth]{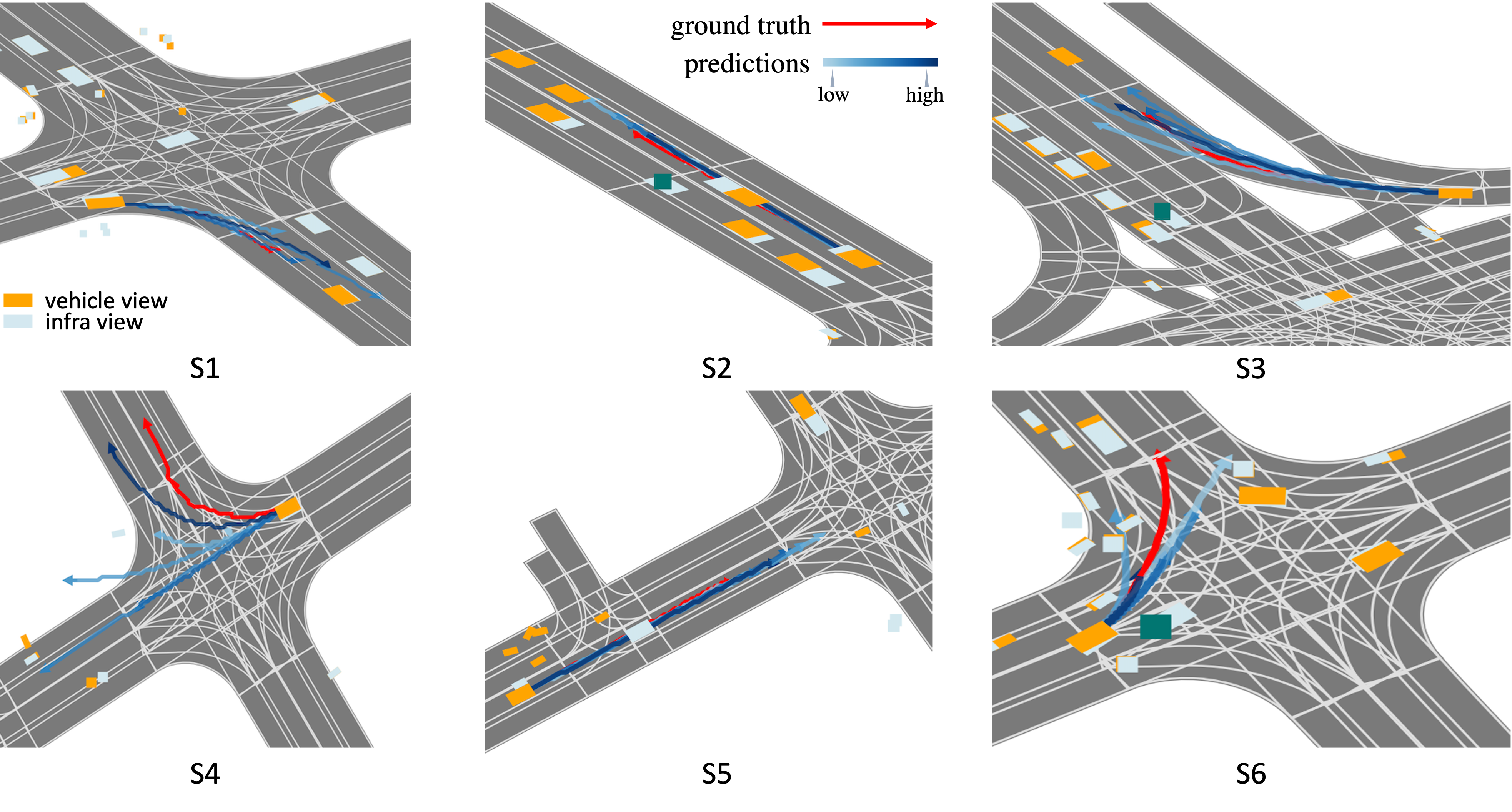}
  \caption[Qualitative results]{Qualitative results. The ground truth (in red) and predicted multimodal trajectories (in different shades of blue) of the target agent are shown. Darker blue represents higher probability. Yellow and grey rectangles denotes road agents observed from vehicle view and infrastructure view, respectively.}\label{figchap5:qualitative}
\end{figure}

\textbf{Qualitative results}. We visualize our prediction results and six representative scenarios are shown in Figure~\ref{figchap5:qualitative}. To maintain clarity and simplicity in the visualization, each scenario displays the ground truth and multimodal prediction results only for the target agent, although predictions for all agents are accessible. In the first column, scenarios S1 and S4 depict the target agent turning right. In the second column, the target agent is shown going straight, leaving (S2) and approaching (S5) an intersection. The last column presents scenarios where the target agent is merging (S3) and turning left (S6). It's shown that our model predictions capture the multimodal behavior. In scenarios S2 and S5, it's exhibited in the form of different velocity profiles. Since lane information is integrated via an attention mechanism rather than as rigid constraints, off-road and road rule-violating predictions and may occur, as shown in S4 and S6. Notably, there are more uncertainties when agent making turns, as evidenced by the dispersion among the predictions. Furthermore, it's important to highlight that the prediction with the highest probability does not always align with the ground truth, as demonstrated in S1 and S3, resulting from the inherent model and data uncertainties. 

\begin{figure}
  \centering  
  \includegraphics[width=\textwidth]{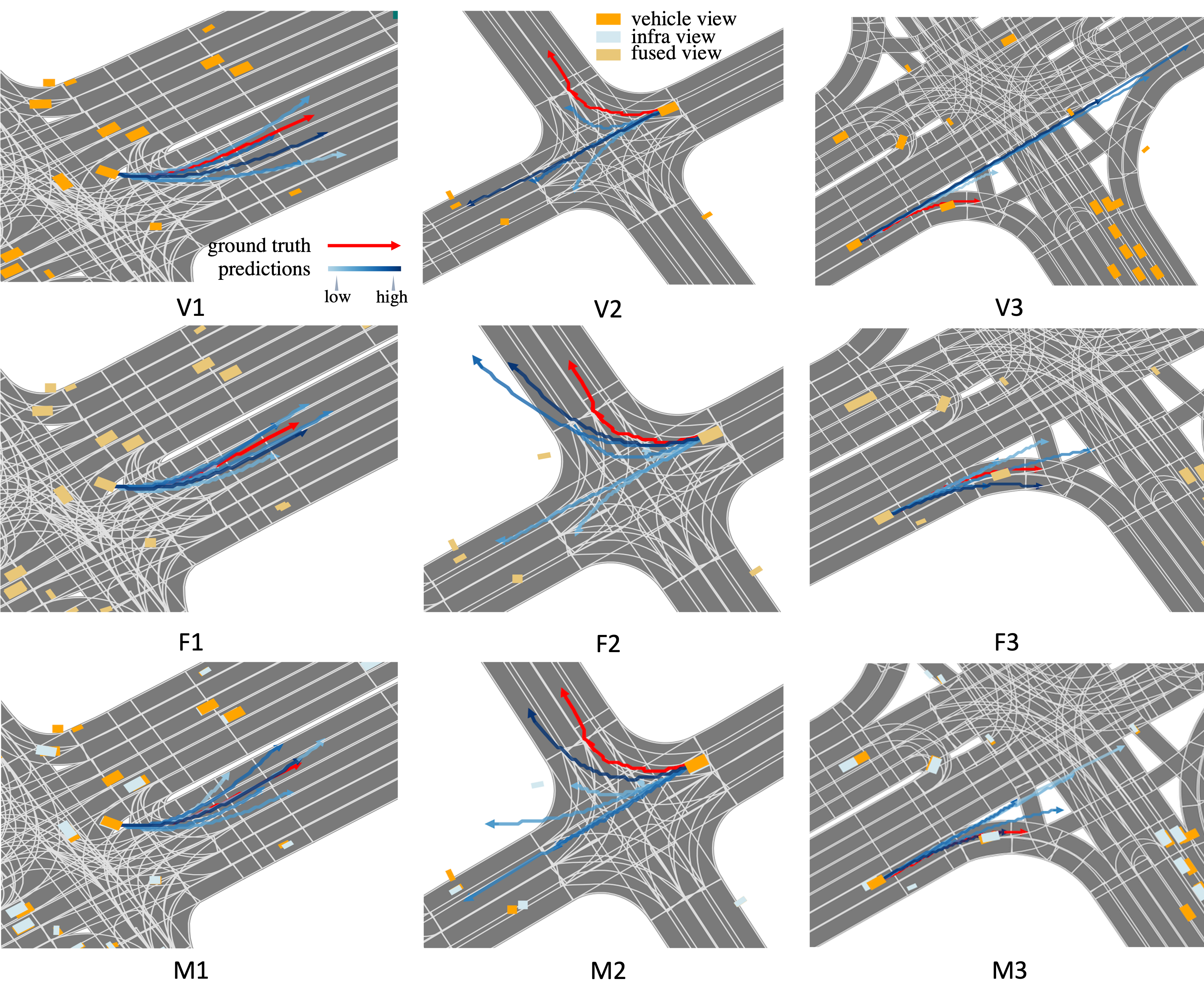}
  \caption[Qualitative results comparison for three models]{Qualitative results comparison for three models: HiVT-Ego (first row), HiVT-PPVIC (second row), and ours (third row). The ground truth (in red) and predicted multimodal trajectories (in different shades of blue) of the target agent are shown. Darker blue represents higher probability. Yellow and grey rectangles denotes road agents observed from vehicle view and infrastructure view, respectively.}\label{figchap5:qualitative2}
\end{figure}

We further compare the prediction results from model HiVT Ego, HiVT PP-VIC and ours on selected scenarios, as shown in Figure~\ref{figchap5:qualitative2}. It's observed that HiVT Ego, relying solely on vehicle-view data, consistently predicted incorrect modes. In contrast, both HiVT PP-VIC and our model, incorporating additional data from the infrastructure view, demonstrate improved accuracy in capturing the ground truth.

\subsubsection{Post-hoc Uncertainty Quantification Performance}
\textbf{Quantitative comparison}. We show in this section that conformal prediction method produces more valid and efficient confidence regions than the model predictions. The evaluation results are shown in Table~\ref{tabchap5:uq}.
\begin{table}[htbp]
\caption{Conformal prediction results comparison}\label{tabchap5:uq}
\centering
\resizebox{\textwidth}{!}{%
\begin{tabular}{|c|c|c|c|c|c|c|c|c|}
\hline
\multirow{2}{*}{Metric} & \multirow{2}{*}{alpha} & \multirow{2}{*}{Mixture} &  \multicolumn{3}{c|}{CF-RNN} & \multicolumn{3}{c|}{CopulaCPTS} \\
\cline{4-9}
 & & &Z-score&L2-norm&L1-norm&Z-score&L2-norm&L1-norm \\
 \hline
 ind. coverage&\multirow{3}{*}{0.2}&0.65&0.99&0.99&0.99&0.93&0.94&0.92 \\
 joint coverage&&0.15&0.97&0.99&0.97&0.76&\textcolor{red}{0.80}&0.76\\
size&&124.43&5734.55&2003.42&432.50&536.09&\textcolor{red}{29.14}&17.83\\
\hline
 ind. coverage&\multirow{3}{*}{0.1}&0.78&0.99&0.99&0.99&0.97&0.96&0.96 \\
 joint coverage&&0.31&0.99&0.99&0.99&0.86&\textcolor{red}{0.93}&0.88\\
size&&215.15&25098.06&3672.24&1138.77&7634.95&\textcolor{red}{64.80}&42.82\\
\hline
 ind. coverage&\multirow{3}{*}{0.05}&0.86&0.99&0.99&0.99&0.97&0.97&0.98 \\
 joint coverage&&0.47&0.99&0.99&0.99&0.99&\textcolor{red}{0.94}&0.95\\
size&&327.48&151062.75&6550.62&2321.75&372768.78&\textcolor{red}{107.63}&190.04\\
\hline
\end{tabular}%
}

\end{table}

The Mixture column presents UQ results derived directly from the multimodal predictions. We first examine the metrics of independent and joint coverage. It's evident that the model predicted intervals fail to meet the validity requirements across all specified miscoverage rates. Interestingly, CF-RNN yields overly conservative results, maintaining coverage levels around 0.99 regardless of the miscoverage rate. This could indicate that Bonferroni correction, often employed with shorter horizons, might not be suitable for our predicted horizon of 50. CopulaCPTS achieves coverage levels very close to the desired values across all specified miscoverage rates $\alpha$.

Next, we inspect the metric "size", which serves as an indicator of UQ efficiency. Benchmark the size from model predictions in the Mixture column, CF-RNN produces extremely large confidence region which is expected considering the overly conservative coverage observed earlier. CopulaCPTS yields comparative and even smaller size compared to the benchmark. 

Finally, we examine the impact of different score functions. In both CF-RNN and CopulaCPTS, Z-score appears as the least efficient among all three functions. Since Z-score is computed using the model-predicted uncertainties $\hat{B}$, it will always be more inefficient than the benchmark size in the "Mixture" column. One plausible explanation provided in \cite{angelopoulos2021gentle}, is that although widely used, $\hat{B}$ in the Z-score function is not directly related to the quantiles of the label distribution. Therefore, it may not be the most suitable score function option in our case. On the other hand, both L2-norm and L1-norm demonstrate significantly higher efficiency, particularly within the CopulaCPTS method. L1-norm, in particular, considers variations among different dimensions, leading to enhanced efficiency. Taking into account both validity and efficiency requirements, we have highlighted the best results among all methods in red.

\begin{figure}
    \begin{subfigure}{\textwidth}
        \centering
        \includegraphics[width=\textwidth]{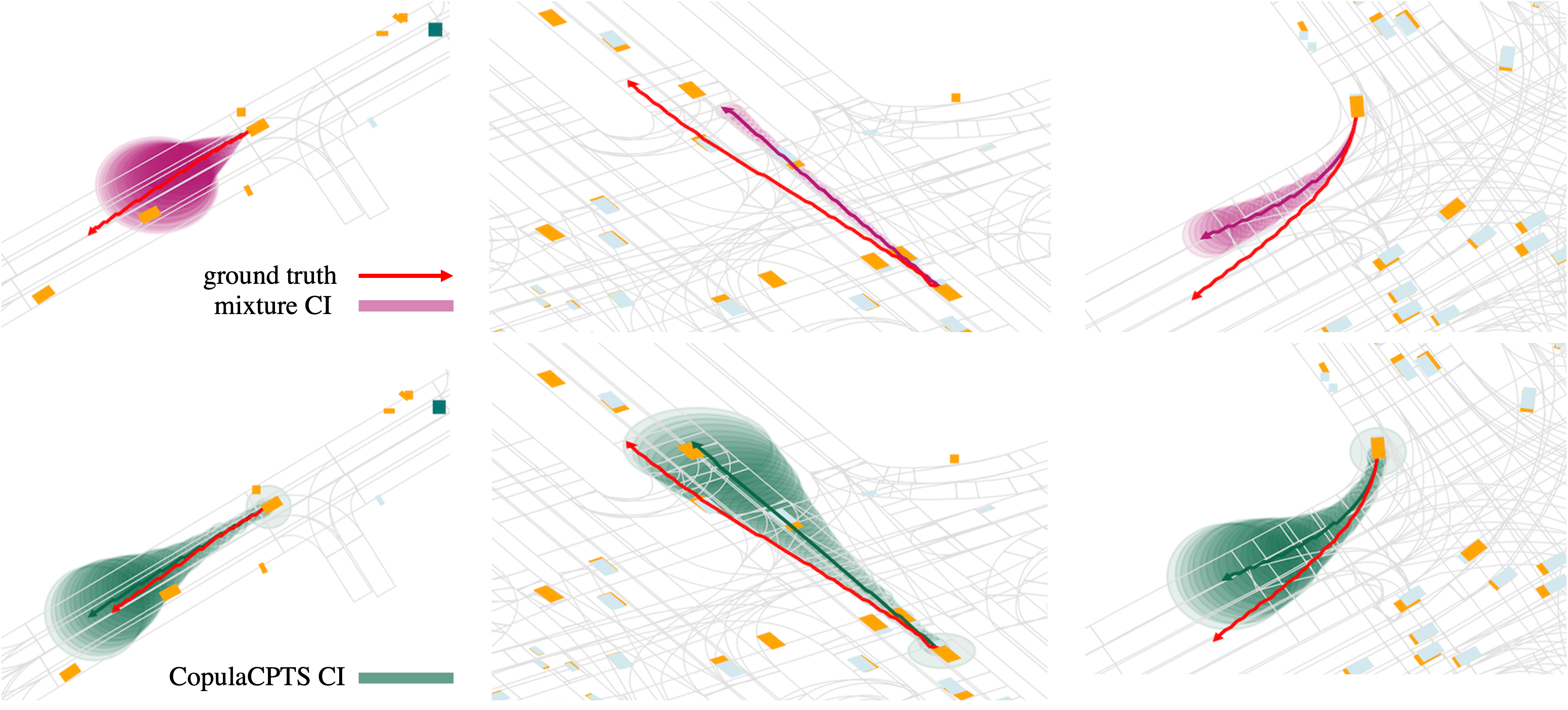}
        \label{figchap5:copula1}
    \end{subfigure}
    \caption[Uncertainty quantification visualization]{Uncertainty quantification comparison for the mixture model prediction (purple shade) and CopulaCPTS (green shade) at $\alpha=0.1$.}
    \label{figchap5: uq}
\end{figure}

\textbf{Qualitative analysis}. We visualize the uncertainty prediction from the mixture model and the CopulaCPTS, as shown in Figure~\ref{figchap5: uq}. Three scenarios are selected in each column. We can see that the mixture model predictions fail to cover the ground truth while CopulaCPTS achieves the coverage at the specified $\alpha$.

\section{Conclusion}
We have presented a novel model framework with multi-view data integration in the cooperative driving setting. Our proposed model is straight forward and can be built upon any existing graph-based single-view models. It has demonstrated its effectiveness and advantages over existing benchmarks. Moreover, we have incorporated a post-hoc uncertainty quantification module, providing valid and efficient confidence regions, which is crucial in safety-critical tasks such as trajectory prediction. 

The proposed framework has certain limitations. From the model's perspective, we currently treat all road agents as the same type. However, in the public V2X-Seq dataset, there are different vehicle types, such as trucks, vans, buses, motorcycles, etc. Future work should address the different characteristics of these vehicle types to enhance model performance. Moreover, better methods for encoding the lane information to eliminate off-road and road-rule-violating predictions should be investigated. From the uncertainty quantification perspective, we simplify the quantile computation on the multimodal prediction into computation on the single best mode. This approach loses valuable distributional information. Therefore, exploring score functions that consider the distribution could lead to more accurate uncertainty quantification for multimodal results. Furthermore, we evaluate the uncertainty assuming independence among the agents. Future work can incorporate agent correlations based on the graph structure to better reflect the underlying uncertainty relations and provide more insightful results.
\bibliographystyle{unsrt}  
\bibliography{references}

\end{document}